# DETECTING MULTIPLE DISEASES IN MULTIPLE CROPS USING DEEP LEARNING

Vivek Yadav[1*] and Anugrah Jain[1]

[1]Computer Science and Engineering, Sardar Vallabhbhai National Institute of Technology, Surat, 395007, Gujarat, India.

*Corresponding author(s). E-mail(s): vivekyadav52454@gmail.com;
Contributing authors: ajain@coed.svnit.ac.in;

**Abstract**

India, as a predominantly agrarian economy, faces significant challenges in agriculture, including substantial crop losses caused by diseases, pests, and environmental stress. Early detection and accurate identification of diseases across different crops are critical for improving yield and ensuring food security. This paper proposes a deep learning based solution for detecting multiple diseases in multiple crops, aimed to cover India's diverse agricultural landscape. We first create a unified dataset encompassing images of 17 different crops and 34 different diseases from various available repositories. Proposed deep learning model is trained on this dataset and outperforms the state-of-the-art in terms of accuracy and the number of crops, diseases covered. We achieve a significant detection accuracy, i.e., 99% for our unified dataset which is 7% more when compared to state-of-the-art handling 14 crops and 26 different diseases only. By improving the number of crops and types of diseases that can be detected, proposed solution aims to provide a better product for Indian farmers.

## 1 Introduction

Agriculture forms the backbone of India's financial system, presenting livelihoods to almost 58% of the population and contributing around 18% to the country's GDP [1]. This sector encompasses diverse crops, including wheat, maize, and horticultural produce. However, crop diseases pose a significant challenge, reducing yields and threatening food security. These diseases, as a result of pathogens consisting of fungi, micro organism, and viruses, can cause annual yield losses predicted between 10% and 30% (FAO, 2022) [2].

Crop diseases significantly threaten agricultural productivity, leading to considerable economic losses. Diseases such as rust in wheat, late blight in potatoes, and powdery mildew in vegetables are prevalent across India. Their early and accurate detection becomes a necessity for improving yield and ensuring food security [3] .

Traditional methods for detection such as visual inspections or clinical diagnosis by specialists take much time and require large human resource. Also, these methods are not scalable for a large agriculture landscape such as in India. Therefore, technologyintensive solutions such as those based on artificial intelligence and machine learning based detection approaches can play a big role for solving them in accurate and scalable manner [4–37].

Various machine learning based detection approached have found in the literature [4–37] They can be categorized into three different categories, i.e., solutions for detecting single disease in single crop [5–8], solutions for detecting multiple diseases in single crop [9–33], and solutions for detecting multiple diseases in multiple crops [34–37]. After an extensive literature survey, we found that they are lacking in terms of coverage for wide range of crops and their diseases. At most, 14 crops and 26 different diseases are covered by the approaches in literature.

This paper proposes a unified dataset based on image collection from various online repositories [38–41]. We cover a total of 17 different crops with a total 34 different types of crop diseases. Proposed deep learning model is trained on this unified dataset and a significant accuracy, i.e., up to 99% is achieved for detecting multiple diseases in multiple crops. We outperform the existing by increasing the accuracy of detection, the number of crops handling, and the number of crop diseases covered.

The rest of the paper is organized as follows. Section 2 presents our literature survey discussing variety of crop diseases and their detection methodologies available in the literature. Section 3 provides details of our unified dataset generated from extraction of images from various available online repositories. Proposed deep learning based detection model is presented in section 4, and section 5 shows our implementation and results. Finally, the paper is concluded in section 6 by providing some pointers to future work.

## 2  Theoretical Background & Literature Survey

In the last few decades, the increase in population and the scarcity of resources have forced some researchers to seek new means for crop monitoring and other crop-related problems, like diseases for instance. There has been extensive utilization of deep learning, which is effective for various tasks, within the identification of crop and plant diseases. However, there are various issues occurred when employing deep learningbased methods such as data imbalance, complex models, less accuracy and/or coverage for multiple number of crops [34–37]. Before discussing all of them one by one, we categorize crops diseases as follows. It will help us to determine what type of diseases are more supported in existing literature.

### 2.1  Crop Diseases

Crop diseases can be categorized following three categories based on the type of pathogen responsible for causing them: (1) fungal, (2) bacterial, and (3) viral. Each category has its own characteristics and impacts on various crops. An overview of each of them is as follows.

1. **Fungal Diseases:** These are common in crops and can cause severe damage. Fungi spread through spores and generally moves with the air, water, and insects. Following are some examples of fungal diseases.

    - Powdery Mildew: Affects crops like wheat, cucumbers, and grapes, resulting in white powdery patches on leaves.
    - Rust Diseases: Generally, seen in wheat and other grains, and are caused by the Puccinia genus, resulting into some reddish-orange pustules on the leaves. • Blight: Named as late blight for tomatoes and potatoes. It is caused by Phytophthora infestans and is popularly known for rapid crop destruction.
    - Downy Mildew: It affects crops like lettuce and grapes, showing yellow patches on leaves with white fungal growth underneath.

2. **Bacterial Diseases:** Bacterial diseases can spread through water, contaminated soil, and tools. They often lead to wilting, leaf spots, and tissue damages. Some example of them are as follows.

    - Bacterial: It is occurred in rice because of Xanthomonas oryzae (a bacteria) and results into the water-soaked leaf spots and yellowing of the leafs.
    - Bacterial Wilt: It is occurred due to a bacteria named as Ralstonia solanacearum, and impacts plants like tomatoes and potatoes. The result is a wilting or a death of the plant.
    - Citrus Canker: Affects citrus trees and cause lesions on leaves, stems, and fruits. • Fire Blight: It impacts apples and pears and mostley resulting into the blackened branches that resemble fire damage.

3. **Viral Diseases:** Viral infections are transmitted by insects, contaminated tools, or through seed. These diseases often lead to stunted growth, mottled leaves, and abnormal plant development. Some of their are examples are listed below. • Tobacco Mosaic Virus (TMV): It affects crops like tobacco, tomatoes and peppers, and cause mosaic-like leaf discoloration.
    - Tomato Leaf Curl Virus (TLCV): It leads to curled and yellowed leaves, and reduces yields in tomatoes.
    - Cucumber Mosaic Virus (CMV): Common in cucumbers and other vegetables, resulting in yellowing and reduced plant vigor.
    - Banana Bunchy Top Virus (BBTV): It affects bananas and cause their stunted growth and bunchy leaves.

## 2.2 Detection Methodologies

Now, we discuss existing deep learning based solutions detecting crop diseases in literature. We classify them in three different categories based on their capability to identify single disease in single crop, multiple diseases in single crop, multiple diseases in multiple crops. Solutions according to all of these categories are discussed below.

### 2.2.1 Detection of Single Disease in Single Crop

Many research papers explore the use of technology to identify specific disease in specific crop. In the study [5], a method was developed to identify corn kernels. The researchers used a face recognition RCNN network for adaptive learning and combined it with color images to detect diseases in grayscale images. This approach has proven robust by yielding results across a wide range of strains and sizes. Another important contribution is presented in [6], whereby a CNN based model for agricultural disease detection was proposed. It has been shown that CNN-based feature extraction outperforms traditional methods such as Wavelet transform with local binary pattern histogram (LBPH) (Haar-WT). Also, classifiers such as Softmax and SVM have achieved good results in disease classification when used with CNNs.

In order to address the issues of finding areas of images and limited data, [7] proposed a new method to detect northern leaves in corn. This approach incorporates CNNs to classify specific images, then another CNN to analyze the entire image, thus improving the classification accuracy. In [8], researchers studied the detection of ginkgo disease using workplace and laboratory data. They used two CNN architectures: VGG16 and InceptionV3, and found that the model's performance varied depending on the information utilized. Together, these research exhibit the ability of CNNs to achieve accuracy and efficiency in crop disease detection.

### 2.2.2 Detection of Multiple Diseases in Single Crop

The deep learning approaches identifying different diseases in certain crop are presented here. In one study [9], a tool for classifying crop diseases was introduced using four deep learning methods. A network was developed and data was collected from various sites and laboratories. The approach presented in [10] uses the MobileNet CNN architecture to identify orange juice diseases based on data generated with the help of agricultural experts. In [11], four distinct diseases were identified using an approach that combine image processing, image and coloration feature extraction, and SVM-primarily based classification.

In another study [12], a transfer learning based approach to detect the Camellia oleifera virus based on AlexNet architecture is presented. Their model achieved a greater accuracy when compared with models such as SVM, AlexNet, ResNet20 and VGG16 [12]. In [13], a deconvolution guided VGG network was proposed to identify and segment tomato leaf disease, addressing the issues of shadow, low light and occlusion. Likewise, [14] developed a unique method to detect tomato leaf spot disease in the material by modifying AlexNet to multi-AlexNet to improve the efficiency.

| Citation | Plants | Disease | Dataset | ModelsforDeepLearning | Accuracy |
|---|---|---|---|---|---|
| 2019 [5] | Wheat | Fusarium Head Blight | Self | Mask-RCNN (CNN) | 92% |
| 2019 [6] | Rice | Rice blast | Self | CNN's new | 95% |
| 2017 [7] | Corn | Northern leaf blight | Self | CNN (AlexNet, GoogLeNet Inception) | 96% |
| 2020 [8] | Ginkgo Biloba | Ginkgo leaf | Self | InceptionV3(CNN),VGGNet-16 | 93% |
| 2020 [9] | Tomato | 6 Diseases | Self | CNN (VGG 19, VGG 16, InceptionV3(CNN), RestNet | 85% |
| 2022 [10] | Apple | Alternaria leaf blotch, rust | Self | MobileNet, RestNet152(CNN) InceptionV3(CNN) | 77% |
| 2020 [11] | Rice | 4 diseases | Self | SVM, New SVM | 96% |
| 2018 [12] | Camellia aleifera | 4 diseases | Self | AlexNet(CNN) | 98% |
| 2020 [13] | Tomato | 10 diseases | Plant Village | Deconvolution Guided VGGNet(CNN) | 93% |
| 2019 [14] | Tomato | 7 diseases | Plant_Village, Self | Improved Multi-Scale AlexNet (CNN) | 92% |
| 2020 [15] | Maize | 9 diseases | Self | R-CNN (Improved Faster R-CNN) | 97% |
| 2019 [16] | Tomato | 5 disease | AI challenger | CNN (New CNN ARNet based) | 89% |
| 2019 [17] | Cucumber | 6 disease | Self | Global pooling dilated(CNN) | 87% |
| 2022 [18] | Grape | 8 disease | AI challenger, Self | Multi-Scale ResNet(CNN) | 88% |
| 2020 [19] | Tomato | 9 disease | Plant-Village | New CNN | 91% |
| 2019 [20] | Apple | 5 disease | Self | INAR SSD | 78% |
| 2020 [21] | Apple | 4 diseases | Plant Pathology 2020 Challenge, Self | ResNet50(CNN) | 97% |
| 2020 [22] | Balsam pear | 5 disease | Self | R-CNN ,Improved Faster R-CNN | 89% |
| 2019 [23] | Wheat | 3 diseases | Self | CNN (ResNet50) | 87% |
| 2024 [24] | Cucumber | 6 diseases | Self | Full CNN | 97% |
| 2022 [25] | Tomato | 14 diseases | Plant | ResNet 34 based Faster RCNN (CNN) | 99% |
| 2023 [26] | Tomato | 3 diseases | Plant Village, Self | Inception V3, Inception ResNet V2(CNN) | 99% |
| 2022 [27] | Bean | 3 diseases | i-Bean | CNN (Mobile-Net, Mobile v2) | 97% |
| 2022 [28] | Corn | 4 diseases | Plant Village | EfficientNet B0 DenseNet 121(CNN) | 98% |
| 2021 [29] | Tomato | 10 diseases | Plant Village | Efficient Net(CNN) | 99% |
| 2022 [30] | Tomato | 6 diseases | PlantVillage | Inception Net(CNN) | 99% |
| 2023 [31] | Tomato | 5 diseases | Self | CNN (PCBAM) | 94% |
| 2022 [32] | Tomato | 10 diseases | PlantVillage | MobileNetV3(CNN) | 98% |
| 2021 [33] | Corn | Leaf blight, rust | Self | CNN(VGG 16) | 95% |
| 2020 [34] | 14 plant species | 26 diseases | Plant Village, IPM, Bing | VGG_16,GoogLeNet, Inception _V3(CNN) | 92% |
| 2020 [35] | Rice, Corn | 9 diseases | PlantVillage, Self | INC-VGGN(CNN) | 92% |
| 2020 [36] | 4 crops | 15 diseases | PlantVillage | Multi-scale ResNet(CNN) | 96% |
| 2019 [37] | 14 crops | 79 diseases | PlantVillage | GoogLeNet | 82% |
| Proposed | 17 crops | 34 diseases | Unified | ResNet(CNN) | 99.02% |

**Table 1**: Comparison of Proposed Approach with State-of-the-art Solutions

To detect the complex background of maize disease in the real world [15] introduced a faster evolution of R-CNN with batch normalization and averaging function, which achieved better accuracy and faster detection. In another study [16], an ARNet architecture combining color and segmentation was proposed for effective classification of tomato diseases. [17] analyzed cucumber leaf disease using Global Pooling Extended CNN (GPD CNN), an enhancement of AlexNet, to address the issue of competition and scale. The multi-scale ResNet CNN in

[18] has demonstrated the accuracy of identifying different fruit diseases. Similarly, [19] introduced a new CNN model that outperformed VGG16, InceptionV3, and MobileNet in tomato leaf disease detection. Instantaneous detection of apple juice was done using INAR-SS method [20], which combines SSD, initial module and rainbow cascade. Another study [21] used ResNet50 CNN network before ImageNet training to classify four juice strains using adversarial contract information. [22] addressed the problem of bitter melon disease by augmenting Faster R-CNN with a pyramid network to solve the variable bounding box size.

A novel ResNet50 architecture that combines image subsampling and reliable data for rice disease detection in regions was introduced in [23]. Cucumber leaf lesion segmentation was improved in [24], which modified the CNN architecture to reduce the learning time while addressing issues such as illumination and future interference. Also, [25] proposed a faster RCNN model that is based on ResNet-34 to recognize a tomato leaf diseases in terms of local time and distribution.

Additional works including [26] and [27] investigate pre-trained methods such as InceptionV3, Inception ResNet V2 and MobileNet for the classification of tomato and bean leaf disease, respectively. [28] introduced the fusion of EfficientNetB0 and DenseNet121 for the classification of maize disease, while [29] combined EfficientNet with modified U-Net for segmented tomato disease. Finally, [30] used semantic segmentation and Inception Net model to identify and segment tomato plants, [31] suggested the parallel convolutional block color module (DIMPCNET) for Dense Inception MobileNet-V2 to solve the problem of identifying tomato leaf diseases in difficult competition. Recent work [32] focused on the detection of tomato diseases using MobileNetV3, which was improved to reduce the over-fitting problem with limited data. In [33], a VGG-16 based model has been presented for corn crop and two disease categories.

### 2.2.3 Detection of Multiple Diseases in Multiple Crops

Recent researches have centered their goal on identifying multiple disease in multiple crops by working on various plant pathogens at the same time. For example, In [34] Suggest a technique utilizing deep learning to classify plant illnesses from photographs. Their strategy implements a CNN model which is modified with plant disease datasets. Through transfer learning, they overcome the challenge associated with limited labeled data, enabling better model performance and effectiveness in accurately diagnosing numerous plant diseases.

Another studies [35] Developed a novel multi-scale vegetable leaf disease identification model based on ResNet. This model utilizes a composite of various scales convolutional filters at different parts of the leaf image which helps the model to capture finer details. With this optimization, detection of subtle signs of disease has improved significantly. Their method further improves recognition accuracy in complex leaf disease patterns by optimizing the ResNet structure.

In [36] concentrated on the identification of plant diseases through deep learning methods from singular lesions and spots on the leaves. The research also puts special emphasis on feature extraction from targeted segments of the leaf using CNNs. Such an approach increases the efficiency of disease diagnosis in plants. This method of detection improves the precision of infection diagnosis by concentrating on visual markers of infection and, thus, improves the precision of diagnosis. In [37], one method involves sectioning leaf lesions and biting them to isolate and diagnose multiple diseases. This method demonstrated its ability to classify one leaf with several diseases.

Although these methods have shown the advantages of deep learning in plant disease detection, they often face issues with coverage, robustness, and scalability. Many specific methods exist for certain crops or limited number of diseases only. The problem of having a generalized model covering large number of diseases in significant number of crops is missing in the literature. Our paper works on this problem and provide an efficient solution outperforming all of these existing methods. In the next section, we discuss how we have collected images from various online repositories to make a unified dataset supporting a good coverage of crops and their occurred diseases.

## 3 Our Unified Dataset

Monitoring and capturing areas of interest visually allow a better analysis of crop growth and its disease identification. Attributes like shape, texture, and color hold significant information. Various existing solutions have used such information to perform crop disease detection [42]. These approaches have used various publicly available image datasets [4–37]. We, in this section, discuss our unified dataset which comprises leaf images taken from various online repositories for covering more number of crops and diseases. After performing this collection, we have performed various data cleaning techniques to make our unified dataset enable for proposed disease classification.

Our unified dataset contains a total of 93136 image samples from crop leaves spanning around 17 different crop species and 34 of their diseases. The dataset includes 51 different images class of healthy conditions as well as diseases, and a total of 34 different diseases are used for unhealthy scenarios. The healthy samples are designated as "Healthy", while the diseased samples are classified according to the type of disease occurred. In this way a total of 51 different categories are generated and each of them are classified using the proposed deep learning model.

Table (2) shows a detailed information of our unified dataset providing the total number of images available in each classification category. Specifically, the training set contains a total of 74651 image samples, and the test and validation set includes 18485 image samples. All images in the dataset are based on RGB model with varying dimensions. Before employing them, we perform their resize operation and all of them are converted into images of 256 × 256 pixels each. By doing this, the computational overhead is minimized and overall efficiency is increased. The finalized dimension and format of the input are 3×256×256 and JPG, respectively. Figure (1) depicts representative samples used as input to the proposed model after performing resize operation. This figure highlights some diversity of collected leaf images and shows their associated conditions. Our dataset collected from various repositories provides a strong basis for performing crop disease detection.

| Crop | Disease | Category | No.ofImages |
|---|---|---|---|
| Apple | Apple Scab | Fungal | 2520 |
| Apple | Apple Black Rot | Fungal | 2484 |
| Apple | Apple Cedar Apple Rust | Fungal | 2200 |
| Apple | Apple Healthy | Healthy | 2510 |
| Banana | Banana Cordana | Fungal | 162 |
| Banana | Banana Healthy | Healthy | 151 |
| Banana | banana Pestalotiopsis | Fungal | 173 |
| Banana | banana Sigatoka | fungal | 473 |
| Blueberry | Blueberry Healthy | Healthy | 2270 |
| Cherry | Cherry (including sour) Healthy | Healthy fungal | 2282 |
| Cherry | Cherry(including sour) Powdery Mildew | | 2104 |
| Corn | Corn(maize) Cercospora Leaf Spot Gray Leaf Spot | Fungal | 2052 |
| Corn | Corn(maize)Common Rust | Fungal | 2384 |
| Corn | Corn(maize) Healthy | Healthy | 2324 |
| Corn | Corn (maize) Northern Leaf Blight | Fungal | 2385 |
| Cotton | Cotton Aphids | Fungal | 700 |
| Cotton | Cotton Army Worm | Fungal | 702 |
| Cotton | Cotton Bacterial Blight | Bacterial | 600 |
| Cotton | Cotton Healthy | Healthy | 700 |
| Cotton | Cotton Powdery Mildew | Fungal | 600 |
| Cotton | Cotton Target spot | Fungal | 601 |
| Grape | Grape Black Rot | Fungal | 2360 |
| Grape | Grape Esca(Black Measles) | fungal | 2400 |
| Grape | Grape Healthy | Healthy | 2115 |
| Grape | Grape Leaf blight(Isariopsis Leaf Spot) | Fungal | 2152 |
| Orange | Orange Haunglongbing(Citrus greening) | Fungal | 2513 |
| Peach | Peach Bacterial spot | Bacterial | 2297 |
| Peach | Peach Healthy | Healthy | 2160 |
| Pepper Bell | Pepper bell Bacterial spot | Bacterial | 2391 |
| Pepper Bell | Pepper bell Healthy | Healthy | 2485 |
| Potato | Potato Early blight | Fungal | 2424 |
| Potato | Potato Healthy | Healthy | 2280 |
| Potato | Potato Late blight | Fungal | 2424 |
| Raspberry | Raspberry Healthy | Healthy | 2226 |
| Soybean | Soybean Healthy | Healthy | 2527 |
| Squash | Squash Powdery mildew | Fungal | 2170 |
| Strawberry | Strawberry Healthy | Healthy | 2280 |
| Strawberry | Strawberry Leaf scorch | Fungal | 2218 |
| Tomato | Tomato Bacterial spot | Bacterial | 2127 |
| Tomato | Tomato Early blight | Fungal | 2400 |
| Tomato | Tomato Healthy | Healthy | 2407 |
| Tomato | Tomato Late Blight | Fungal | 2314 |
| Tomato | Tomato Leaf Mold | Fungal | 2352 |
| Tomato | Tomato Septoria leaf Spot | Fungal | 2181 |
| Tomato | Tomato Spider mites Two-spotted spider mite | Fungal | 2176 |
| Tomato | Tomato Target Spot | Fungal | 2284 |
| Tomato | Tomato Mosaic virus | Virus | 2238 |
| Tomato | Tomato Yellow Leaf Curl Virus | Virus | 2451 |
| Wheat | Wheat Healthy | Healthy | 102 |
| Wheat | Wheat Septoria | Fungal | 97 |
| Wheat | Wheat stripe rust | Fungal | 208 |

**Table 2**: Crops and Diseases in Our Unified Dataset

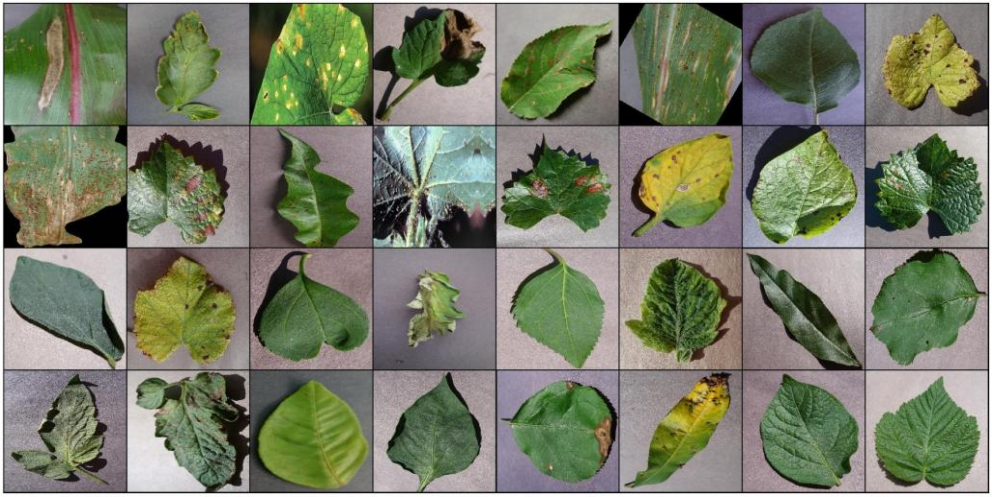

**Fig. 1**: Input Image

## 4 Proposed Detection Methodology

In this section, we discuss architecture of our detection model employing a residual network with convolutional layers. We use an instance of ResNet9 from ResNet family of convolutional neural networks for proposed detection purpose. Compared to deeper architectures like ResNet50 and ResNet101, it is efficient in terms of computational complexity, making it ideal for deployment at places where only mobiles/edge devices are available such as in rural areas of agriculture landscape. We use the concept of residual blocks which are central to ResNet designs. These blocks use shortcut connections to bypass layers, ensuring the network learns residual mappings rather than direct mappings. This approach mitigates issues like vanishing gradients and accelerates training convergence. Figure 2 illustrates our deep learning model, and all of its internal details are discussed step-by-step as follows.

### 4.1 Network Architecture

The residual network architecture is well-suited for crop disease detection, as it balances computational efficiency with high classification accuracy. Leveraging the principles of residual learning, this model is intended to process and extract important features from incoming images of diseased and healthy crop leaves, providing a robust solution for disease classification task while maintaining the less computational requirement. Below is a detailed explanation of our residual neural network architecture.

### 4.1.1 Input Layer

The first layer is the input layer which takes RGB images of crop leaves, earlier resized to a standard dimension of 3×256×256 pixels to make certain consistency throughout the dataset. The size (3, 256, 256) represents height, width, and number of channels used for each input image, respectively.

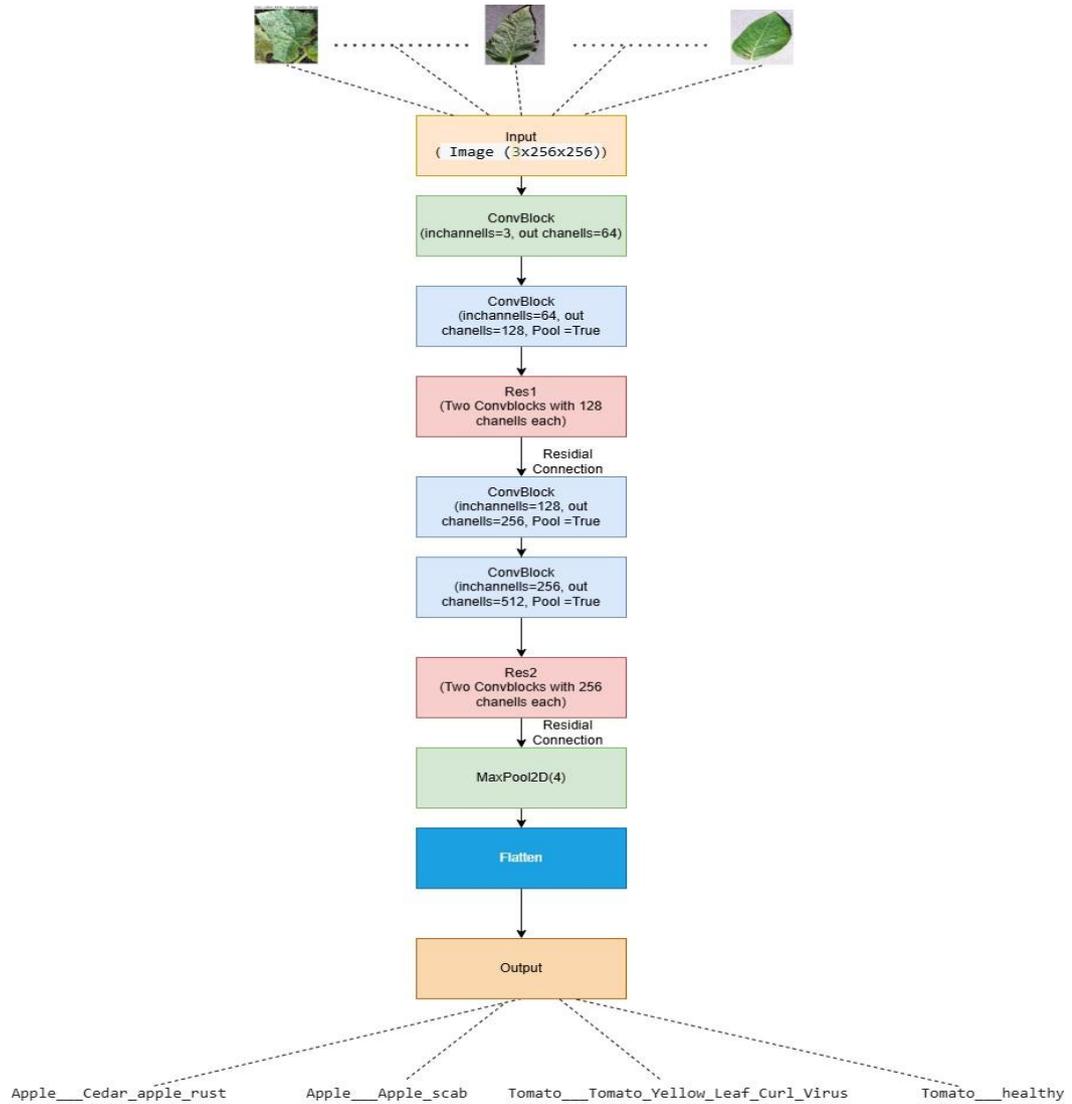

**Fig. 2**: Our Deep Learning Model

### 4.1.2 Convolutional Layers

The feature extraction begins with two initial convolutional layers. The first layer has 64 filters with a kernel size of 3x3, stride of 1, and padding of 1. Low-level characteristics of images such as edges and textures are captured by this layer. The second convolutional layer increases the number of filters to 128 while reducing the spatial resolution by half i.e. stride of 2. This decreases the computational load on the network and enables it to concentrate more on high-level properties. Following each convolutional layer, we have a ReLU activation function to expedite and stabilize training, incorporating batch normalization and non-linearity.

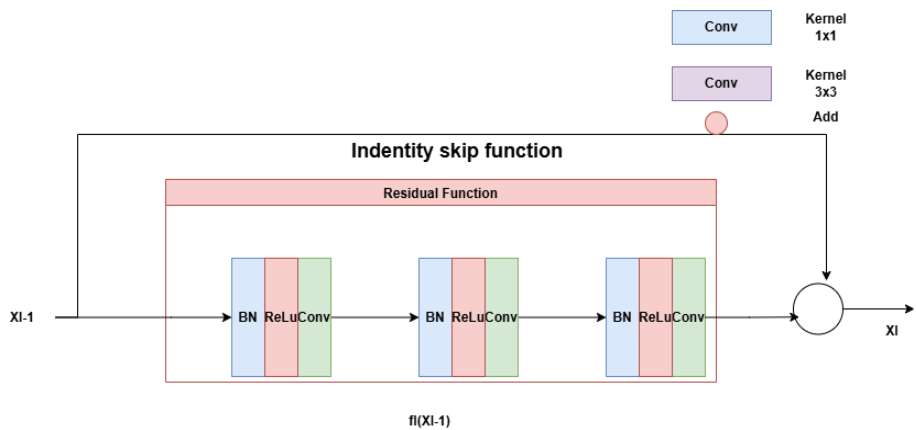

**Fig. 3**: Residual Block

### 4.1.3 Residual Blocks

Our model includes three residual blocks taken from are the standard ResNet architecture. Every residual block as shown in fig. (3) consists of two convolutional layers with identical filter counts and kernel sizes. Batch normalization and ReLU activation functions are used after each convolution. Also, a skip connection that directly adds the block's input to its output is employed ensuring that the gradients can flow freely during back-propagation. By enabling the network to learn residual mappings, these blocks facilitate optimization and enhance the network's capacity to generalize to new input. The skip connections are especially valuable in addressing challenges like vanishing gradients and ensuring deeper layers do not degrade model performance.

### 4.1.4 Max Pooling

After performing feature extraction, the spatial dimensions of the feature maps are reduced by employing max pooling. In order to preserve the most important activations, this process pulls the maximum value from each region of the input feature map. Furthermore, it supports translation invariance, over-fitting management, and reduction of the computational complexity.

| Category | Parameter | Specification |
|---|---|---|
| **Model Configuration** | Hardware | Google Colab T4 TPU |
| | Framework | PyTorch |
| | Dependencies | NumPy, Pandas, Matplotlib |
| | Learning Rate | 0.001 |
| | Scheduler | Cosine Annealing |
| | Optimization Algorithm | Adam |
| | Regularization (Weight Decay) | 0.0001 |
| | Training Iterations | 5 Epochs |
| **Performance Metrics** | Accuracy | |
| | Precision | |
| | Recall | |
| | F1-score | |
| | Confusion Matrix | |
| | | Visual Representation |

**Table 3**: Model Implementation Details

### 4.1.5 Fully Connected Layer

Our fully connected layer serves as the classifier for the detected features. It maps the high-level features extracted by the convolutional and residual layers to a set of probabilities, one for each class of crop disease. [Mention the number of classes in proposed model here]

### 4.1.6 Activation Functions

We use ReLU activation function after each convolutional and batch normalization layer to introduce non-linearity and avoid issues such as vanishing gradients. In the final classification layer, Softmax activation function is used to generate normalized class probabilities.

# 5 Implementation and Results

This section discusses our implementation setup and classification results. We first provide parameters used for model training and validation. Afterwards, the results of the proposed detection methodology for our unified dataset are presented.

## 5.1 Experimental Setup

We have performed our experiments on Google Co-lab and have utilized a TPU (T4) there. Our model is implemented using PyTorch framework with the main programming language as Python 3.7. Various libraries are involved including Pandas, NumPy, and Matplotlib for handling different processes and visualization. The model was setup for training with initial learning rate of 0.001 and a batch length of 32. A weight decay parameter of 0.0001 was used with Adam optimizer. To avoid over-fitting issue, 5 epochs for training were used.

Various evaluation metrics have used to compare proposed detection methodology with state-of-the-art approaches such as accuracy, recall, precision, and F1-score. Table (3) provides a summary of model configuration and evaluation parameters. In the next subsection, we provide a brief introduction to our evaluation parameters.

## 5.2 Evaluation Metrics

We use four well-known evaluation parameters for proposed detection. Our main goal is to assess its capacity to precisely categorize diseases. Following are a brief introduction to them.

1. **Accuracy:** Accuracy represents the general proportion of correct predictions made by the model. It is calculated by considering both the correctly identified positive and negative instances.
   **Formula:**
   $$\text{Accuracy} = \frac{TP + TN}{TP + TN + FP + FN} \qquad (1)$$

2. **Precision:** Precision quantifies the proportion of positive cases that were really positive. It indicates the reliability of the positive predictions made by the model.
   **Formula:**
   $$\text{Precision} = \frac{TP}{TP + FP} \qquad (2)$$

3. **Recall:** It's another name is sensitivity which quantifies the percentage of real positive examples that the model accurately detected. It shows how well the model can detect relevant instances.

**Formula:**

$$\text{Recall} = \frac{TP}{TP + FN} \quad (3)$$

4. **F1-Score:** Precision and Recall's harmonic mean is the F1 Score. It combines both metrics into one score, It is especially helpful when there is an uneven distribution of classes and a need is there to balance recall and precision.
   **Formula:**

$$\text{F1 Score} = 2 \cdot \frac{\text{Precision} \cdot \text{Recall}}{\text{Precision} + \text{Recall}} \quad (4)$$

## 5.3 Results

This section presents results of the proposed detection methodology classifying multiple diseases in multiple crops efficiently. Our results are identified for each category of crops and its diseases. Four evaluation parameters as discussed in above section are used and their final values after performing complete experimentation are presented in Table (4).

Our approach demonstrates exceptional performance for detecting multiple diseases in multiple crops. For some disease it is performing exceptionally well such as for detecting Potato-Earlyblight, Squash-Powderymildew, Cherry-Powderymildew, and Peach-healthy, it achieves a perfect score for all the parameters. The classes corresponding to healthy crops consistently yield high accuracy and F1-scores. For example, F1-scores for Raspberry-healthy, Strawberry-healthy, Corn-healthy are 0.9955, 0.9989, and 0.9968, respectively. This indicates robustness of model in distinguishing healthy crops from diseased ones.

| Class | Accuracy | Recall | Precision | F1 Score |
|---|---|---|---|---|
| Apple - Scab | 0.998 | 0.9882 | 0.998 | 0.9931 |
| Apple - Black rot | 1 | 0.998 | 1 | 0.999 |
| Apple - Cedar rust | 1 | 0.9977 | 1 | 0.9989 |
| Apple - healthy | 0.998 | 0.9882 | 0.998 | 0.9931 |
| Blueberry - healthy | 1 | 0.9913 | 1 | 0.9956 |
| Cherry(including sour) - Powdery mildew | 1 | 1 | 1 | 1 |
| Cherry (including sour) - healthy | 0.9978 | 0.9956 | 0.9978 | 0.9967 |
| Corn(maize) - Cercospora leaf spot Gray leaf spot | 0.961 | 0.9875 | 0.961 | 0.974 |
| Corn(maize) - Common rust | 0.9979 | 0.9979 | 0.9979 | 0.9979 |
| Corn(maize) - Northern Leaf Blight | 0.9895 | 0.9652 | 0.9895 | 0.9772 |
| Corn(maize) - healthy | 0.9978 | 0.9957 | 0.9978 | 0.9968 |
| Grape - Black rot | 0.9936 | 1 | 0.9936 | 0.9968 |
| Grape - Esca (Black Measles) | 1 | 0.9979 | 1 | 0.999 |
| Grape - Leaf blight (Isariopsis Leaf Spot) | 1 | 1 | 1 | 1 |
| Grape - healthy | 1 | 1 | 1 | 1 |
| Orange - Haunglongbing (Citrus greening) | 0.998 | 0.996 | 0.998 | 0.997 |
| Peach - Bacterial spot | 0.9891 | 1 | 0.9891 | 0.9945 |
| Peach - healthy | 0.9977 | 0.9977 | 0.9977 | 0.9977 |
| Pepper, bell Bacterial spot | 0.9979 | 0.9958 | 0.9979 | 0.9969 |
| Pepper - bell healthy | 0.992 | 0.998 | 0.992 | 0.995 |

| Class | Accuracy | Recall | Precision | F1 Score |
|---|---|---|---|---|
| Potato - Early blight | 1 | 1 | 1 | 1 |
| Potato - Late blight | 0.9897 | 0.9938 | 0.9897 | 0.9917 |
| Potato - healthy | 0.9956 | 0.9956 | 0.9956 | 0.9956 |
| Raspberry - healthy | 0.9933 | 0.9977 | 0.9933 | 0.9955 |
| Soybean - healthy | 0.996 | 1 | 0.996 | 0.998 |
| Squash - Powdery mildew | 0.9977 | 1 | 0.9977 | 0.9988 |
| Strawberry - Leaf scorch | 0.991 | 1 | 0.991 | 0.9955 |
| Strawberry - healthy | 1 | 0.9978 | 1 | 0.9989 |
| Tomato - Bacterial spot | 0.9835 | 0.9905 | 0.9835 | 0.987 |
| Tomato - Early blight | 0.9771 | 0.971 | 0.9771 | 0.974 |
| Tomato - Late blight | 0.9741 | 0.9762 | 0.9741 | 0.9751 |
| Tomato - Leaf Mold | 0.9979 | 0.9958 | 0.9979 | 0.9968 |
| Tomato - Septoria leaf spot | 0.9794 | 0.9839 | 0.9794 | 0.9816 |
| Tomato - Spider mites Two-spotted spider mite | 0.9724 | 0.9953 | 0.9724 | 0.9837 |
| Tomato - Target Spot | 0.9737 | 0.9591 | 0.9737 | 0.9663 |
| Tomato - Yellow Leaf Curl Virus | 0.9939 | 1 | 0.9939 | 0.9969 |
| Tomato - mosaic virus | 0.9978 | 0.9955 | 0.9978 | 0.9967 |
| Tomato - healthy | 0.9979 | 0.9877 | 0.9979 | 0.9928 |
| Wheat - Healthy | 0.0909 | 0.2222 | 0.0909 | 0.129 |
| Wheat - Septoria | 1 | 0.4857 | 1 | 0.6538 |
| Wheat - stripe rust | 0.8163 | 0.9524 | 0.8163 | 0.8791 |
| Banana - cordana | 0.9697 | 0.9412 | 0.9697 | 0.9552 |
| Banana - healthy | 0.9091 | 0.8696 | 0.9091 | 0.8889 |
| Banana - pestalotiopsis | 0.875 | 0.875 | 0.875 | 0.875 |
| Banana - sigatoka | 0.9764 | 0.992 | 0.9764 | 0.9841 |
| Cotton - Aphids | 0.99 | 0.99 | 0.99 | 0.99 |
| Cotton - Army worm | 1 | 0.9444 | 1 | 0.9714 |
| Cotton - Bacterial Blight | 0.95 | 1 | 0.95 | 0.9744 |
| Cotton - Healthy | 1 | 0.9901 | 1 | 0.995 |
| Cotton - Powdery Mildew | 0.98 | 0.9899 | 0.98 | 0.9849 |
| Cotton - Target spot | 0.9505 | 0.932 | 0.9505 | 0.9412 |
| **Overall Performance** | **0.9903** | **0.9903** | **0.9903** | **0.9901** |

**Table 4**: Our Classification Results Showing Accuracy, Recall, Precision and F1 Score

Despite the complexity of multiple class classification, we maintain a higher accuracy for most of the crops including widely used Apple, Grape, Corn, and Potato. Many classes have well-balanced precision, recall, and F1-scores, representing that our model has successfully reduced false positives and false negatives.

The overall value (99.03%) of recall indicates model's ability in identifying actual positive cases. The overall precision has been measured as 99.03%, demonstrating how well the provided model predicts positive classes. The F1 score which is calculated as 99.01% demonstrates the close alignment of these metrics and a favorable balance between recall and precision of the model, in turn highlighting the resilience and consistent performance of the model even in the case of some imbalanced data. Our all findings indicate that the presented model is best suited for multiple diseases detection and outperforms the state-of-the-art in multiple crops and multiple diseases detection category.

# 6 Conclusion

This paper presents a deep learning model for detecting multiple diseases in multiple crops efficiently. Traditional methods such as manual inspections by experts are unable to scale for the large agriculture landscape of India. Deep learning based solutions as presented in the literature are easier to use, deploy, and scalable in nature. However, there is no single solution covering all of the mostly used crops and their diseases in India. Our paper tries to construct such a dataset by merging information, i.e., crop leaf images from various available online repositories so that a generalized detection model can be designed and implemented. We cover 17 different crops and 31 different diseases in our dataset which is maximum among we found in the literature. Our deep learning model which is based on residual networks has trained on this dataset. Compared to state-of-the-art, we achieve a significant 99.03% accuracy which is maximum in the category of multiple diseases for multiple crops classification. In future, we will extend this dataset to include more number of crops such as rice and perform a better classification with proposed or some new deep learning models.

## Declarations

**Funding** Not Applicable.
**Competing Interests** The author declares no conflict of interest.
**Author contribution** All the authors have contributed equally.
**Data Availability Statement** Data will be made available on reasonable request.
**Research Involving Human and /or Animals** Not Applicable.
**Informed Consent** Not Applicable.